\documentclass[journal, a4, 11pt, twoside, twocolumn]{IEEEtran}

\usepackage{newtxmath}
\usepackage{amsmath}
\usepackage{amsfonts}
\usepackage{mathtools, nccmath}

\usepackage[superscript, noadjust]{cite}

\usepackage{graphicx}
\graphicspath{{figures/}}
\DeclareGraphicsExtensions{.pdf,.jpeg,.png,.eps}

\usepackage{booktabs}
\usepackage{array}
\usepackage{footnote}
\usepackage[acronym]{glossaries}
\usepackage{color}
\usepackage{enumitem}
\usepackage{stfloats}
\usepackage[hidelinks]{hyperref}
\usepackage[protrusion, expansion, kerning, spacing]{microtype}

\usepackage{siunitx}
\newacronym{snn}{SNN}{Spiking Neural Network}
\newacronym{dnn}{DNN}{Deep Neural Network}
\newacronym{stdp}{STDP}{Spike-Timing Dependent Plasticity}
\newacronym{bptt}{BPTT}{Back-Propagation Through Time}

\title{Micro-power spoken keyword spotting on Xylo Audio 2}
\author{Hannah Bos$^{1}$, Dylan R. Muir$^{1,*}$ %
    \thanks{$^1$SynSense, Zürich, Switzerland. $^*$Correspondence to Dylan R. Muir, \href{mailto:dylan.muir@synsense.ai}{dylan.muir@synsense.ai}. }%
}
\date{June 2024}

\begin{document}

\maketitle

\begin{abstract}
    For many years, designs for ``Neuromorphic'' or brain-like processors have been motivated by achieving extreme energy efficiency, compared with von-Neumann and tensor processor devices.
    As part of their design language, Neuromorphic processors take advantage of weight, parameter, state and activity sparsity.
    In the extreme case, neural networks based on these principles mimic the sparse activity oof biological nervous systems, in ``Spiking Neural Networks'' (SNNs).
    Few benchmarks are available for Neuromorphic processors, that have been implemented for a range of Neuromorphic \textit{and} non-Neuromorphic platforms, which can therefore demonstrate the energy benefits of Neuromorphic processor designs.
    Here we describes the implementation of a spoken audio keyword-spotting (KWS) benchmark ``Aloha'' on the Xylo Audio 2 (SYNS61210) Neuromorphic processor device.
    We obtained high deployed quantized task accuracy, (95\%), exceeding the benchmark task accuracy.
    We measured real continuous power of the deployed application on Xylo.
    We obtained best-in-class dynamic inference power (\SI{291}{\micro\watt}) and best-in-class inference efficiency (\SI{6.6}{\micro\joule\per Inf}).
    Xylo sets a new minimum power for the Aloha KWS benchmark, and highlights the extreme energy efficiency achievable with Neuromorphic processor designs.
    Our results show that Neuromorphic designs are well-suited for real-time near- and in-sensor processing on edge devices.
\end{abstract}


Xylo™ Audio is a family of ultra-low-power audio inference chips, designed for in- and near-microphone analysis of audio in real-time energy-constrained scenarios.
Xylo is designed around a highly efficient integer-logic processor which simulates parameter- and activity-sparse spiking neural networks (SNNs) using a leaky integrate-and-fire (LIF) neuron model.
Neurons on Xylo are quantised integer devices operating in synchronous digital CMOS, with neuron and synapse state quantised to \SI{16}{\bit}, and weight parameters quantised to \SI{8}{\bit}.
Xylo is tailored for real-time streaming operation, as opposed to accelerated-time operation in the case of an inference accelerator.
Xylo Audio includes a low-power audio encoding interface for direct connection to a microphone, designed for sparse encoding of incident audio for further processing by the inference core.

In this report we present the results of a spoken KWS audio benchmark deployed to Xylo Audio 2.
We describe the benchmark dataset; the audio preprocessing approach; and the network architecture and training approach.
We present the performance of the trained models, and the results of power and latency measurements performed on the Xylo Audio 2 development kit.
We include for comparison previous benchmarks of the Aloha KWS task on other neuromorphic devices, mobile inference processors, CPUs and GPUs.

\section*{Benchmark dataset}

\begin{figure}
    \centering
    \includegraphics[width=44.5mm]{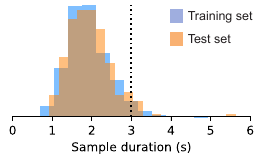}
    \caption{
        \textbf{Distribution of sample durations in the Aloha dataset.}
        In this work we pad and clip samples to a uniform \SI{3}{\second} duration (dashed line).
        This retains the majority of data in both train and test datasets.
    }
    \label{fig:dataset_distribution}
\end{figure}

We implemented the ``Aloha'' benchmark dataset introduced by Blouw et. al\cite{blouw_benchmarking_2019}.
This includes a training set of approximately 2000 utterances from 96 speakers, with a 3:1 ratio between the target phrase (``aloha'') and non-target phrases.
Figure~\ref{fig:dataset_distribution} shows the distribution of sample durations in the training and test datasets.
In the results below we use the provided test set of 192 samples.

We designed a network that identified ``aloha'' target samples by producing one or more output events, and non-target samples by remaining silent.
We clipped or extended samples to \SI{3}{\second} by padding with silence.

\section*{Audio preprocessing}

\begin{figure}
    \centering
    \includegraphics[width=\linewidth]{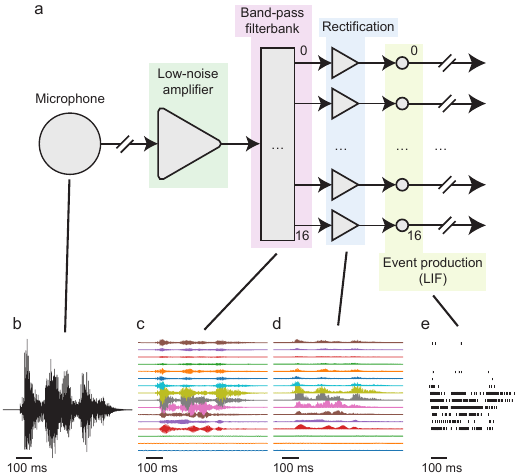}
    \caption{
        \textbf{Audio preprocessing approach.}
        \textbf{a} The stages of audio preprocessing in Xylo Audio 2.
        Single-channel audio arrives at a microphone (\textbf{b}).
        This passes through a band-pass Butterworth filterbank, and is split into $N=16$ frequency bands (\textbf{c}).
        Filter output is rectified (\textbf{d}) before passing through a bank of LIF neurons that smooth and quantize the signals in each band.
        The result is a set of sparse event channels (\textbf{e}), where the firing intensity in each channel is proportional to the instantaneous energy in each frequency band.
    }
    \label{fig:audio_preprocessing}
\end{figure}

We encoded each sample as sparse events, using a simulation of the audio encoding hardware present on the Xylo Audio device.
The design of this preprocessing block is shown in Figure~\ref{fig:audio_preprocessing}\cite{muir_rockpool_2023}.
Briefly, this block is a streaming-mode buffer-free encoder, designed to operate continuously on incoming audio.
A low-noise amplifier with a selectable gain of \num{0}, \num{6} or \SI{12}{\deci\bel} amplifies the incoming audio.
A band-pass filter bank with 2nd-order Butterworth filters splits the signal into 16 bands, with centre frequencies spanning \SIrange{40}{16940}{\hertz} and with a Q of 4.
The output of these filters is rectified, then passed through a leaky integrate-and-fire (LIF) neuron to smooth the signal and convert it to events.
The result is to convert a single audio channel into 16 sparse event channels with event rate in each channel corresponding to the energy in each frequency band.

Samples were trimmed to \SI{3}{\second}, encoded using the preprocessing block described here, and binned temporally to \SI{100}{\milli\second}.
Our approach operates in streaming-mode, analysing a continuous time-frequency representation of the input audio, similar to a real-time Fourier transform (see Figure~\ref{fig:audio_preprocessing}e).

\section*{Network architecture}

\begin{figure*}
    \centering
    \includegraphics[width=\textwidth]{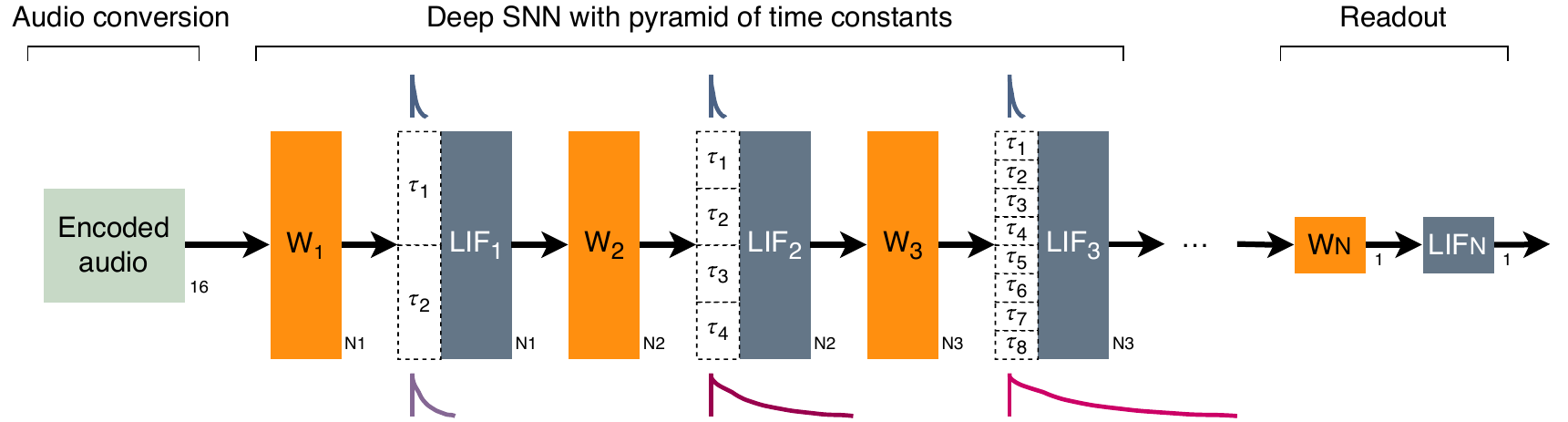}
    \caption{
        \textbf{The SynNet architecture used in this benchmark.}
        Event-encoded audio is provided as input, as described in Figure~\ref{fig:audio_preprocessing}.
        The network consists of a single feed-forward chain of fully-connected layers, using the LIF neuron model.
        Several time constants are distributed over each layer, with shorter time constants in early layers and longer time constants in later layers (see text for details).
        A single readout LIF neuron is used in each network.
    }
    \label{fig:synnet_architecture}
\end{figure*}

We use a feed-forward spiking neural network architecture called ``SynNet''\cite{bos_sub-mw_2022} (Figure~\ref{fig:synnet_architecture}).
This is a fully-connected multi-layer architecture, interleaving linear weight matrices with LIF neuron layers.
Each layer has a number of synaptic time constants, where the time constants are defined as $\tau_n = 2^n * \SI{10}{\milli\second}$, and neurons are evenly distributed with the range of time constants for that layer.
Layers with two time constants therefore have half the neurons with synaptic time constants $\tau_1 = \SI{20}{\milli\second}$ and half with $\tau_2 = \SI{40}{\milli\second}$.
Layers with 4 time constants have a quarter of the neurons with synaptic time constants $\tau_1 = \SI{20}{\milli\second}$; a quarter with $\tau_2 = \SI{40}{\milli\second}$; a quarter with $\tau_3 = \SI{80}{\milli\second}$ and so on.
Membrane time constants for all neurons, as well as readout neuron time constants, are set as $\tau_m = \SI{20}{\milli\second}$.

We describe a given SynNet network architecture in the following by defining the list of hidden layer widths $H$ and corresponding list of numbers of time constants $\tau$.
For example, the network $H = \left[160, 60, 60, 60, 60, 60 \right]$ $\tau = \left[2, 2, 4, 4, 8, 8\right]$ has 6 hidden layers with a first hidden layer width of 160 neurons, followed by 60 neurons, and so on; and with the first hidden layer containing 2 synaptic time constants, the second with 2 synaptic time constants, the third with 4 and so on.
One readout LIF neuron is present in each network, designed to be active for the target class (the keyword ``aloha'') and inactive for any non-target audio.

\section*{Training}

Networks were defined using the open-source Rockpool toolchain (\url{https://rockpool.ai}), with the \textit{torch} back-end.
During training, the membrane potential of the readout neuron was taken as the network output.
Targets were defined as $y=0$ for a non-target sample and $y=1$ for a target sample.

The training loss for readout channels was defined as follows.
\begin{align*}
    \textrm{PeakLoss}(\textrm{x}, y) & =
    \begin{cases}
        \textrm{MSE}\left(1/M\int_m^{m+M}\textrm{x}, \textrm{g}\right) \textrm{ if } y = 1 \\
        w_l \cdot \textrm{MSE}\left(\textrm{x}, \textbf{0}\right) \textrm{ if } y = 0
    \end{cases}
\end{align*}
where
$\textrm{x}$ is a membrane potential vector over time for a single readout channel;
$y$ is the target for the channel (either $1$ indicating a target for this channel in this sample, or $0$ indicating a non-target for this sample);
MSE is the mean-squared-error loss function;
$m = \arg\max\textbf{x}$ is the index of the peak value in $\textbf{x}$;
$M$ is the window duration to examine from $\textbf{x}$ following the peak;
$\textbf{g}$ is a vector $g\cdot \textbf{1}$, a target value that $\textbf{x}$ should match around its peak;
$w_l$ is a weighting for the non-target loss component;
$\textbf{0}$ is the vector of all-zeros.
For the networks trained here, we took $M=\SI{140}{\milli\second}$, $g = 1.5$ and $w_l = 1.4$.

Models were trained for 300 epochs, using the PyTorch Lightning package to manage training.

\section*{Model performance}

Several trained models with a range of total model sizes were evaluated for task performance on the test set.
The presence of any readout events during a sample was taken as a ``target'' prediction.
We computed the true positive and false positive rates over the test set, as well as the binary accuracy.
The model performance for several model sizes is shown in Table~\ref{tab:model_performance}.

\begin{table}
    \resizebox{\linewidth}{!}{%
    \begin{tabular}{rllll}
        $N_{tot}$ & Model & Acc. & TPR & FPR \\ \hline
        461 & $H=[160, 60, 60, 60, 60, 60]$ & 96.88\% & 97.92\% & 4.17\% \\
        411 & $H=[110, 60, 60, 60, 60, 60]$ & 97.92\% & 97.92\% & 2.08\% \\
        401 & $H=[150, 50, 50, 50, 50, 50]$ & 97.40\% & 93.75\% & 6.25\% \\
        361 & $H=[60, 60, 60, 60, 60, 60]$  & 94.27\% & 97.92\% & 8.33\% \\
        341 & $H=[140, 40, 40, 40, 40, 40]$ & 98.44\% & 100.0\% & 3.12\% \\
        281 & $H=[130, 30, 30, 30, 30, 30]$ & 93.23\% & 94.79\% & 8.33\% \\
        221 & $H=[120, 20, 20, 20, 20, 20]$ & 97.40\% & 95.83\% & 1.04\% \\
        $^\dagger$461 & $H=[160, 60, 60, 60, 60, 60]$ & 95.31\% & 91.67\% & 1.04\% \\
        \hline \\
    \end{tabular}
    }
    \caption{
        \textbf{Test performance for trained models.}
        Time constants $\tau = [2, 2, 4, 4, 8, 8]$ for all models.
        $N_{tot}$: Total neurons in the model;
        Acc.: Accuracy;
        TPR: True Positive Rate $TPR = TP / (TP + FN)$;
        FPR: False Positive Rate $FPR = FP / (FP + TN)$.
        $^\dagger$Quantised model result deployed to the Xylo architecture.
    }
    \label{tab:model_performance}
\end{table}



\begin{figure}
    \centering
    \includegraphics[width=\linewidth]{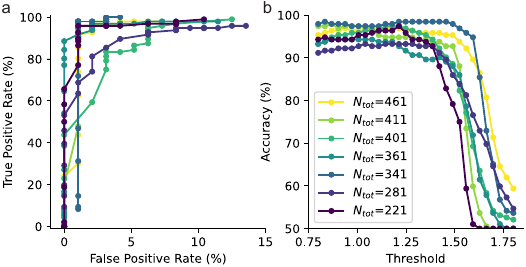}
    \caption{
        \textbf{ROC curves for the trained models in Table~\ref{tab:model_performance}.}
        \textbf{a} True Postive Rate vs False Positve Rate curves.
        \textbf{b} Accuracy for the several models while varying the threshold of the readout neuron.
    }
    \label{fig:roc_curves}
\end{figure}

We computed ROC (Receiver Operator Characteristic) curves for the trained models on the test set, by varying the threshold of the output neuron (Figure~\ref{fig:roc_curves}).

\section*{Power and inference rate}

\begin{figure}
    \centering
    \includegraphics[width=60mm]{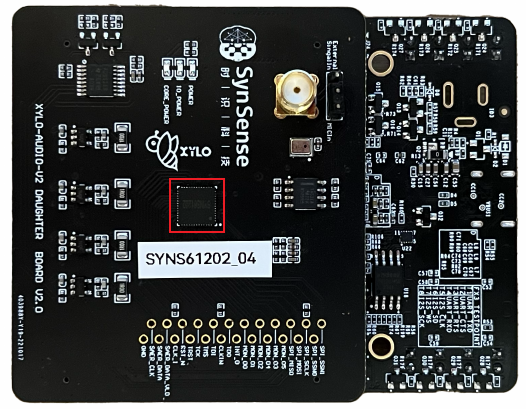}
    \caption{
        \textbf{The Xylo™ Audio 2 hardware development kit (HDK).}
        The HDK is a USB bus-power board requiring a PC-host for power and interfacing.
        The HDK interfaces with the open-source Rockpool toolchain for deployment and testing.
        An analog microphone and a analog jack are provided for direct analog single-channel differential input.
        Encoded audio data can alternatively be streamed from the host PC.
        Inference is performed on the Xylo device (red outline).
    }
    \label{fig:xyloa2_hdk}
\end{figure}

Models were quantized and deployed to Xylo devices using the Rockpool deployment pipline.
Power was measured on the benchmark application deployed to a Xylo Audio 2 device, on the Xylo Audio 2 hardware development kit (Figure~\ref{fig:xyloa2_hdk}), during streaming continuous analysis of the Aloha test-set.
The master clock frequency for Xylo Audio was set to \SI{6.25}{\mega\hertz}.
Current measurements were taken using on-board current monitors, at a frequency of \SI{1280}{\hertz}.
Active power was measured while streaming encoded audio for the entire test set to the Xylo device.
Idle power was measured by deploying a model to the device, then measuring consumed power for five seconds with no model input.
Inference rate was defined in line with Blouw et al., with one inference corresponding to the processing of 10 time-steps by the network\cite{blouw_benchmarking_2019}.

For all trained models, Xylo required idle power of \SIrange{216}{217}{\micro\watt} and active power of \SIrange{468}{514}{\micro\watt}, resulting in dynamic power of \SIrange{251}{298}{\micro\watt}.
Inference rate varied with model size, ranging \SIrange{40}{102}{Inf\per\second}.
This corresponds to dynamic energy per inference of \SIrange{2.4}{7.3}{\micro\joule\per Inf}.
Considering active energy per inference, we computed a range of \SIrange{4.6}{12.7}{\micro\joule\per Inf}.

\section*{Comparison with other benchmark measurements}

\begin{table}
    \resizebox{\linewidth}{!}{%
    \begin{tabular}{llllll}
        Hardware & Idle & Act.& Dyn. & Dyn. E  & Act. E  \\
         &  (\SI{}{\milli\watt}) & (\SI{}{\milli\watt}) & (\SI{}{\milli\watt}) & (\SI{}{\milli\joule\per Inf}) & (\SI{}{\milli\joule\per Inf}) \\
        \hline
        GPU\cite{blouw_benchmarking_2019} & 14970 & 37830 & 22860 & 29.67 & 49.1 \\
        CPU\cite{blouw_benchmarking_2019} & 17010 & 28480 & 11470 & 6.32 & 15.7 \\
        Jetson\cite{blouw_benchmarking_2019} & 2640 & 4980 & 2340 & 5.58 & 11.9 \\
        MOVIDIUS\cite{blouw_benchmarking_2019} & 210 & 647 & 437 & 1.5 & 2.2 \\
        LOIHI\cite{blouw_benchmarking_2019} & 29 & 110 & 81 & 0.27 & 0.37 \\
        LOIHI\cite{yan_comparing_2021} & 29 & 40 & 11 & 0.037 & 0.13 \\
        SpiNNaker2\cite{yan_comparing_2021} & --- & --- & 7.1 & 0.0071 & --- \\
        \textbf{Xylo (ours)} & \textbf{0.216} & \textbf{0.507} & \textbf{0.291} & \textbf{0.0066} & \textbf{0.011} \\
        \hline \\
    \end{tabular}
    }
    \caption{
        KWS task energy benchmarking in comparison with traditional and neuromorphic architectures.
        Power measured on physical devices in all cases.
        Act.: Active;
        Dyn.: Dynamic;
        E: Energy per inference.
        Active energy is not reported for SpiNNaker2 in the source benchmark paper\cite{yan_comparing_2021}, but this is elsewhere reported as \SI{390}{\milli\watt}\cite{nazeer_language_2024}.
    }
    \label{tab:energy-per-inference}
\end{table}

\begin{figure}
    \centering
    \includegraphics[width=\linewidth]{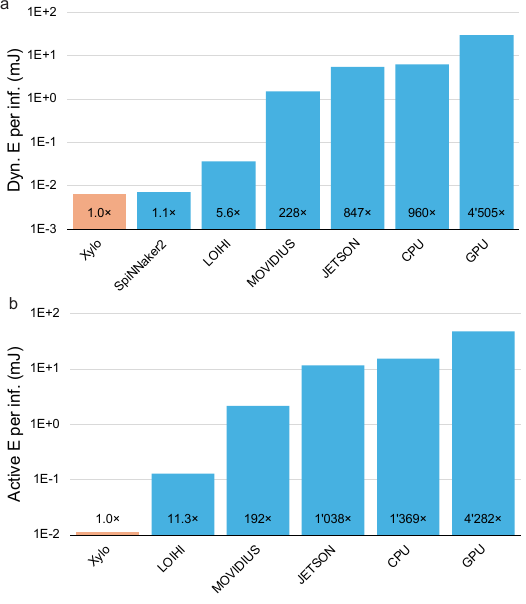}
    \caption{
        \textbf{Energy per inference comparison for the Aloha KWS benchmark task on several hardware architectures.}
        \textbf{a} Dynamic energy per inference comparison.
        This is the standard metric reported for the Aloha benchmark.
        \textbf{b} Active energy per inference comparison.
        Active energy is not reported for SpiNNaker2 in the source benchmark paper\cite{yan_comparing_2021}, but this is elsewhere reported as \SI{390}{\milli\watt}\cite{nazeer_language_2024}.
        Energy per inference for Xylo defined as baseline (\SI{1.0}{\times}).
        See Table~\ref{tab:energy-per-inference} for precise values.
    }
    \label{fig:energy_per_inf}
\end{figure}

We compared our results with several other hardware deployments of the same benchmark task (Table~\ref{tab:energy-per-inference}; Figure~\ref{fig:energy_per_inf}).
The model deployed to Xylo Audio exhibited the lowest continuous idle, active and dynamic power consumption of any of the comparison devices.
Previous results for the Aloha benchmark report dynamic energy per inference; Xylo Audio achieved the lowest dynamic energy per inference of all comparison devices.

For the devices where total active power was reported, we also compared active energy required per inference, as we believe this is a more realistic system-level metric.
Xylo Audio achieved the lowest active energy per inference by an order of magnitude (Table~\ref{tab:energy-per-inference} Act. E).

\section*{Discussion}
We implemented the Aloha spoken KWS benchmark task on Xylo Audio 2.
Our trained network achieved high task accuracy despite its compact size, and the deployed quantised network suffered from only a small drop in accuracy (\SI{<2}{\percent}).
We measured power used by the physical Xylo Audio 2 device while performing inference on the benchmark test set, and computed the inference rate for the system.

We found that Xylo Audio 2 exhibited high task accuracy (higher than the benchmark standard of 93\%); performed inference faster than real-time (\SI{>4}{\times} speedup); and required \SI{291}{\micro\watt} of dynamic power for inference.
Xylo Audio 2 beat all other benchmarked devices on idle power, active power, dynamic power and inference efficiency.

The benchmark results reported here, as well as reported benchmarks for other hardware devices, do not include the power required for audio preprocessing.
Most implementations of the Aloha benchmark require computation of an MFCC spectrogram, which can be computationally demanding.
We used a simulation of the Xylo Audio 2 audio encoding block for audio preprocessing in simulation (Figure~\ref{fig:audio_preprocessing}).
We have measured the power consumed by the audio pre-processing block on Xylo Audio 2 as \SI{<50}{\micro\watt}.

Xylo Audio 2 is designed to operate as a real-time device for in- and near-sensor signal processing.
Here we are operating the device in accelerated time, achieving a speed-up of \SI{>4}{\times}, and an inference rate of \SI{>40}{\hertz}.
This is a lower inference rate than obtained for inference accelerator designs such as LOIHI, SpiNNaker2 and GPUs.
These devices are designed to operate at high inference rates on large volumes of data, often making extensive use of parallel processing.
In contrast with these systems, Xylo is designed to be an efficient real-time processor, operating on a continuous (i.e. non-batched) real-world signal.
This is reflected by the energy efficient performance of Xylo at moderate inference rates.


Our results underscore the efficiency of Neuromorphic processor designs.
Previous benchmark results for other Neuromorphic devices have shown large-factor gains in energy efficiency over low-power conventional processors\cite{rutishauser_7_2023}, mobile inference processors and inference ASICs\cite{frenkel_reckon_2022, parpart_implementing_2023, ottati_spike_2024}, commodity CPUs\cite{blouw_benchmarking_2019, ostrau_benchmarking_2022, parpart_implementing_2023, kosters_benchmarking_2023} and GPUs\cite{blouw_benchmarking_2019, ostrau_benchmarking_2022, parpart_implementing_2023, kosters_benchmarking_2023}.
We show that Xylo Audio 2 sets a new standard for general-purpose Neuromorphic processors, exhibiting micro-power operation on continuous real-time signal processing tasks.


\bibliographystyle{plain}
\bibliography{bibliography.bib}

\end{document}